\documentclass[11pt,british]{IEEEtran}

\usepackage[T1]{fontenc}
\usepackage[utf8]{inputenc}
\usepackage[a4paper]{geometry}
\usepackage{fancyhdr}
\usepackage{babel}
\usepackage{array}
\usepackage{float}
\usepackage{calc}
\usepackage{url}
\usepackage{graphicx}
\usepackage{tablefootnote}
\usepackage{setspace}
\usepackage[unicode=true,pdfusetitle,
 bookmarks=true,bookmarksnumbered=false,bookmarksopen=true,bookmarksopenlevel=1,
 breaklinks=false,pdfborder={0 0 1},backref=false,colorlinks=false]
 {hyperref}

\makeatletter

\providecommand{\tabularnewline}{\\}

\usepackage{fancyhdr}
\makeatother

\usepackage[style=authoryear]{biblatex}
\begin{document}
\title{Using Unsupervised Learning to Help Discover the Causal Graph}
\author{Seamus (Simon James) Brady,{\small{}}\\
{\small{}Dept. of Electronic and Computer Engineering, UL.}}
\maketitle
\begin{abstract}
The software outlined in this paper, AitiaExplorer, is an exploratory
causal analysis tool which uses unsupervised learning for feature
selection in order to expedite causal discovery. In this paper the
problem space of causality is briefly described and an overview of
related research is provided. A problem statement and requirements
for the software are outlined. The key requirements in the implementation,
the key design decisions and the actual implementation of AitiaExplorer
are discussed. Finally, this implementation is evaluated in terms
of the problem statement and requirements outlined earlier. It is
found that AitiaExplorer meets these requirements and is a useful
exploratory causal analysis tool that automatically selects subsets
of important features from a dataset and creates causal graph candidates
for review based on these features. The software is available at \url{https://github.com/corvideon/aitiaexplorer}
\end{abstract}

\section{Introduction}

Causality has become a major research interest in the field of machine
learning. Although collecting vast amounts of data leads to the creation
of useful systems, in order to truly understand the data one needs
to understand the causes and effects that are implicit in the data.
As Peters et al. (2017) point out,\footnote{\textcite{petersElementsCausalInference2017}, page 5.}
causal reasoning is actually ``more powerful'' than just using probabilistic
and statistical reasoning alone. Unlocking causality will allow our
machine learning systems to do more with the data available.

The software outlined in this paper, AitiaExplorer (from the Ancient
Greek for cause, \textit{aitía}), directly addresses this need to
unlock causality within our systems. AitiaExplorer is an exploratory
causal analysis (ECA)\footnote{\textcite{mccrackenExploratoryCausalAnalysis2016}.}
tool which uses unsupervised learning for feature selection in order
to expedite causal discovery. Exploratory causal analysis is an initial
first step in the causal analysis of a system, a causal counterpart
to exploratory data analysis\footnote{\textcite{behrensPrinciplesProceduresExploratory1997}.}.
ECA can be at least partially automated and the results can guide
more formal causal analysis.

Feature selection methods are used in data analysis to suggest subsets
of features that reduce dimensionality while minimising information
loss. These methods can be divided into standard (or ``classical'')
feature selection methods and causal feature selection methods. Classical
feature selection methods identify subsets of features using non-causal
correlations. Causal feature selection methods seek to identify subsets
of features by capturing actual casual relationships between selected
features. However, \textcite{yuUnifiedViewCausal2018} argue that
both ``causal and non-causal feature selection have the same objective''.
This research suggests that even though classical feature selection
methods have a different methodology to causal feature selection methods,
they both work by exploiting common underlying structures in the data.\footnote{Both methods leverage the ``parent and child'' relationships between
features, i.e. Markov blankets. See \textcite{yuUnifiedViewCausal2018}.} Unfortunately, causal feature selection methods are less computationally
efficient than classical feature selection methods and causal feature
selection is still an active area of research.\footnote{For an overview see \textcite{yuCausalitybasedFeatureSelection2019}.}
Classical feature selection is much better understood and the algorithms
are generally faster. This is why AitiaExplorer provides an easy way
to exploit standard feature selection methods as part of exploratory
causal analysis (ECA) rather than causal feature selection methods.

One of the main outputs of ECA are causal graphs. Causal graphs\footnote{Please see later sections for more technical discussion on causal
related terms.} are a useful tool in exploring causality but creating the correct
causal graph is difficult. AitiaExplorer facilitates straightforward
causal exploration of even quite large datasets by automatically creating
causal graphs of subsets of the features of the dataset.

\begin{figure}[H]
\begin{centering}
\includegraphics[scale=0.22]{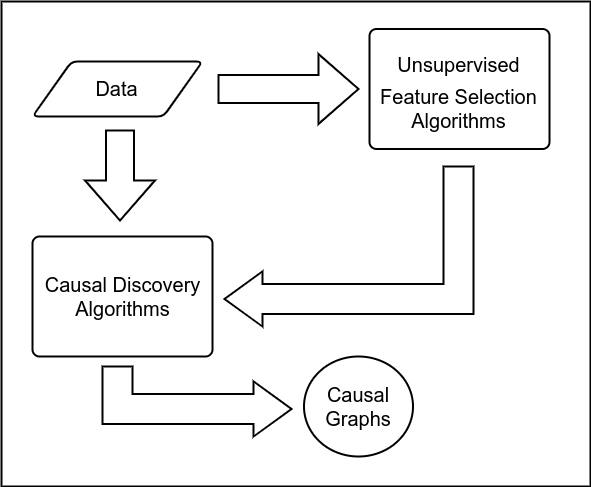}
\par\end{centering}
\textbf{\caption{System overview.}
}
\end{figure}

This paper makes the following contributions:
\begin{itemize}
\item An overview and evaluation of the AitiaExplorer system which contains
an ensemble of both causal discovery algorithms (responsible for creating
the causal graphs) and an ensemble of unsupervised feature selection
algorithms (responsible for automatically selecting the most important
features). The two ensembles work together as outlined in \textbf{Figure
1}.
\item The use of clustering analysis for feature selection via Principal
Feature Analysis.
\item The innovative use of standard supervised learning algorithms to allow
their use in unsupervised feature selection. This is achieved by using
synthetic data generated via a Bayesian Gaussian Mixture model.
\item The ability of AitiaExplorer to create causal graphs from a dataset
when no known target causal graph is provided.
\item The ability of AitiaExplorer to automatically provide insights into
latent unobserved variables in a causal graph.
\end{itemize}
This paper is structured as follows: In Section 2, the problem space
of causality is described and a concise problem statement is provided.
In Section 3, an overview of related research is provided. In Section
4, key requirements in the implementation of AitiaExplorer are discussed
and in Section 5 the key design decisions behind AitiaExplorer are
explored. In Section 6, the actual implementation of AitiaExplorer
is described and in Section 7, this implementation is evaluated in
terms of the problem statement and requirements outlined earlier.

\section{Background}

\subsection{What is Causation?}

The dictionary definition of causation\footnote{The author notes the cliché that ``correlation is not causation''
and leaves it to one side.} (or more correctly \textit{causality}) is ``the act or agency which
produces an effect''\footnote{https://www.merriam-webster.com/dictionary/causation}
or the ``connection between two events or states such that one produces
or brings about the other''.\footnote{http://www.businessdictionary.com/definition/causality.html}
This is certainly close to the intuitive, idiomatic idea of causation
that one uses in daily life. An action produces an effect. For example,
the rain \textit{caused} the grass to be wet. The act of walking in
a puddle without shoes \textit{causes} my feet to get wet.

Philosophers have argued over the metaphysics of causality for nearly
two millennia and modern psychology has added more layers of ambiguity
to these discussions.\footnote{\textcite{whiteIdeasCausationPhilosophy1990}.}
This level of discussion is very interesting but a simpler definition
will suffice for this article. The definition used in this document
is that given by Pearl et al. (2016):
\begin{quote}
``A variable \textit{X} is a \textit{cause} for a variable \textit{Y}
if \textit{Y} in any way relies on \textit{X} for its value''. \footnote{\textcite{pearlCausalInferenceStatistics2016}, page 5.}
\end{quote}
This definition will cover intuitive everyday ideas of causality (for
example, the \textit{grass} relies on the \textit{rain} for its wetness)
and will also cover causal graphs later, where \textit{X} and \textit{Y}
will express variables in a more formal causal inference.\footnote{Please note that the term \textit{causal inference} is defined in
a more exact manner below.}

\subsection{Causation and the Sciences}

To many in the sciences during the early twentieth century, there
was ``a general suspicion of causal notions {[}which{]} pervaded
a number of fields outside of philosophy, such as statistics and psychology''.\footnote{\textcite{galethompsonCausationPhilosophyScience2006}.}
However, despite efforts to banish causality from statistical research
in fields such as medicine, many scientists continued to attempt to
answer causal questions, despite having inadequate statistical training.
As Hernan et al. put it, even today ``confusions generated by a century-old
refusal to tackle causal questions explicitly are widespread in scientific
research''.\footnote{\textcite{hernanSecondChanceGet2019}.} However,
causality has become a mainstream concern for scientists. In the words
of one author, now it seems that we are all becoming social scientists\footnote{\textcite{grimmerWeAreAll2015}.}
as the causal analysis techniques that were widely used in social
and biological research are moving into other fields such as machine
learning.

\subsection{Competing Methods of Learning Causality}

Unfortunately, for a student of machine learning entering the field
of causality, there still seems to be no general agreement on what
way is appropriate and correct for approaching causal inference. Depending
on the field, whether it be social science, biological science or
statistics, there are overlapping definitions and competing claims
on what is most important. In a review of the available methods of
learning causality, Guo et al. (2018) point out that ``... the two
most important formal frameworks ... {[}are{]} ... the \textit{structural
causal models} and the \textit{potential outcome} framework{[}s{]}...''
(emphasis added)\footnote{\textcite{guoSurveyLearningCausality2018}.}.
However, if we consult another review of the available methods in
Lattimore and Ong (2018), we find that they break down the competing
schools of causality into \textit{counterfactuals (Neyman--Rubin)},\textit{
Structural Equation Models} and \textit{causal Bayesian networks},
with a short mention of \textit{Granger causality}. As we can see,
even in two recent review papers of the field, there are multiple
competing naming conventions\footnote{See \textcite{lattimorePrimerCausalAnalysis2018}.}. 

\subsection{Causal Model Framework (CFM)}

\begin{figure}
\includegraphics[scale=0.28]{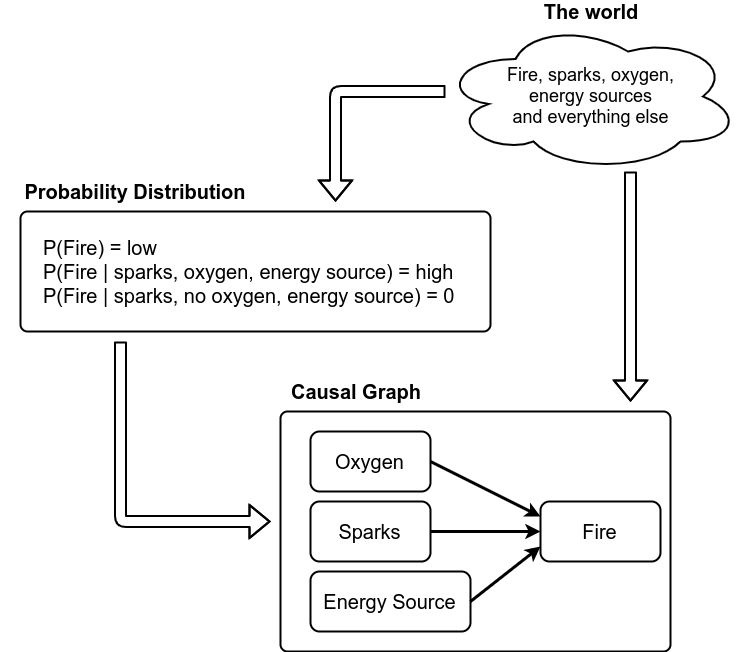}

\caption{Causal Model Framework.}
\end{figure}

This document will use what Steven Sloman calls the ``causal model
framework''.\footnote{\textcite{slomanCausalModelsHow2009}, page 36.}
This framework unites many of the disparate causal elements identified
above into one scheme. He points out that this model is first described
in 1993\footnote{via Bayesian networks in \textcite{spirtesCausationPredictionSearch1993}.}
and then completed in the work of Judea Pearl\footnote{In particular see \textcite{pearlCausalityModelsReasoning2000}.}.
The causal model framework consists of three entities:
\begin{enumerate}
\item A representation of the state of a system. This generally means some
kind of dataset which acts as a proxy for the system.
\item A probability distribution that describes the causal connections in
the dataset.
\item A causal graph depicting this dataset.\footnote{\textcite{slomanCausalModelsHow2009}, page 37.}
\end{enumerate}
From this description we can see that any useful causal discovery
tool needs to support these three entities.

The major advantage of using the causal model framework (CMF) is that
using this scheme bypasses much of the confusing causal terminology
that is used across several dozen scientific fields from genomics
to social science. CMF allows a student of machine learning to use
Bayesian networks, Structural Equation Models (SEM)\footnote{\textcite{bollenStructuralEquationModels2005}.}
and even potential outcomes\footnote{\textcite{imbensRubinCausalModel2010}.}
(which are normally seen as an alternative, if not competition to
SEM) in modelling a causal system. What brings all of these methods
together in CMF is the causal graph. See \textbf{Figure 2} for an
illustration of CMF.\footnote{Adapted from \textcite{slomanCausalModelsHow2009}, page 39.}

\subsection{What is a Causal Graph?}

In general, causal graphs are graphical models that are used to provide
formal and transparent representation of causal assumptions. They
can be used for inference and testing assumptions about the underlying
data. They can also be known as path diagrams or causal Bayesian networks.

More formally in CMF, a causal graph or \textit{graphical causal model}
consists of a directed acyclic graph (DAG) where an edge between nodes
$X$ and $Y$ represents a function $f_{Y}$ that assigns a value
to $Y$ based on the value of $X$. We can say in this case that $X$
is a cause of $Y$. In general, parent nodes are causes of child nodes.
For a quick and thorough look at the relationship between all these
terms see \textit{An Introduction to Causal Inference} by Pearl.\footnote{\textcite{pearlIntroductionCausalInference2010}.} 

For instance, in the example graph in \textbf{Figure 3}, one can see
that the edge between nodes $sprinkler$, $rain$ and $wet$ represents
a function $f_{wet}$ that assigns a value to $wet$ based on the
value of $rain$ and $sprinkler$. You can see that $rain$ and $sprinkler$
are both causes of $wet$.

\begin{figure}
\begin{centering}
\includegraphics[scale=0.33]{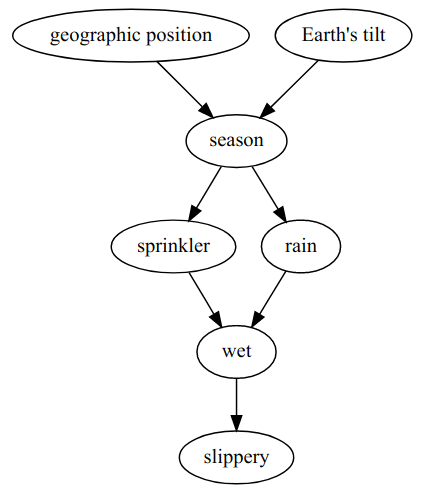}
\par\end{centering}
\caption{Example causal graph.}
\end{figure}

\subsection{Causal Inference, Causal Discovery and Learning Causality}

The causal model (also known as a \emph{Structural Causal Model})
underlying the causal graph is the conceptual model that describes
the causal mechanisms of the system, i.e. the functions represented
above by the edges of the causal graph. In CMF, the causal model is
captured in the probability distributions that describe the underlying
causal connections in the data.

But a causal model can be expressed with or without or a causal graph.
For example, a set of Structural Equation Models (SEM)\footnote{The relationship between causal models and SEMs is not clear in the
literature. See \textcite{bollenEightMythsCausality2013} for more
details.} can describe the causal mechanisms of a system without a causal graph.

If one has a causal graph \textit{and} a causal model for a system,
this means that one has a comprehensive overview of the causal structure
of that system. Guo et al. (2018) call the process of finding the
causal graph \textit{causal discovery}\footnote{The various methods for causal discovery in AitiaExplorer are discussed
later in this paper. } and the process of working with the causal model \textit{causal inference}.
Both of these terms come under the more general term of \textit{learning
causality}. There are other ways of defining these terms of course,
but in general this is a useful distinction to make as it allows for
more clarity in refining the research problem of this document. 

\subsection{How a Causal Discovery Algorithm Works}

To understand the causal model framework and learning causality in
a little more depth, it is useful to consider the logic behind one
of the first causal discovery algorithms\footnote{First described in \textcite{vermaEquivalenceSynthesisCausal1991}.},
the IC Algorithm (Inductive Causation).

The relationship between a causal model and the real world system
is the joint probability distribution over the variables in the system.
Within the causal model framework, there are two assumptions made
concerning this relationship:
\begin{enumerate}
\item The \emph{causal Markov condition}\footnote{See\emph{ }\textcite{slomanCausalModelsHow2009}, page 47.}
which assumes that the direct causes of a variable, i.e. its parents,
make it probabilistically independent of other variables in the system.
Once one knows the values of the parental causes, there is no need
to go back though long chains of indirect causes to find the value
of a variable.
\item The \emph{stability} or \emph{faithfulness} \emph{assumption}\footnote{Ibid.}
which assumes that the probabilistic independencies captured in the
causal graph are because of an underlying causal structure and not
just randomness.
\end{enumerate}
One of the outcomes of these assumptions is the criterion of \emph{d-separation}
(where the \emph{d }stands for \emph{dependence}). Consider three
sets of nodes in a causal graph $X$, $Y$ and $Z$. Set $Z$ is said
to d-separate the nodes in $X$ from the nodes in $Y$ if and only
if $Z$ blocks all paths (edges), and hence information, from a node
in $X$ to a node in $Y$.\footnote{Adapted from \textcite{pearlCausalityModelsReasoning2000}, page 17.}

Consider the example graph from \textbf{Figure 3}:
\begin{itemize}
\item \textit{geographic position} and \textit{Earth's tilt} are causally
independent. 
\item \textit{season}, \emph{sprinkler}, \emph{rain} and \emph{wet} are
causally dependent on \textit{geographic position} and \textit{Earth's
tilt}. 
\item \textit{geographic position} and \textit{Earth's tilt} are direct
causes of \textit{season }\textit{\emph{(an example of a }}$v$-structure).
\item \textit{geographic position} and \textit{Earth's tilt} are indirect
causes of \emph{sprinkler}, \emph{rain} and \emph{wet}. 
\item If \textit{season} is fixed, for example to \textquotedbl Winter\textquotedbl ,
i.e. changes are prevented to \textit{geographic position} and \textit{Earth's
tilt}, then \textit{geographic position} and \textit{Earth's tilt}
will no longer cause changes in \emph{sprinkler}, \emph{rain} and
\emph{wet}. This is also called \emph{blocking} or \emph{controlling}
for the \textit{season }\textit{\emph{node.}}
\item Node \emph{sprinkler} can be said to be d-separated (and hence conditionally
independent) from node \emph{rain} when we control for the \textit{season
}\textit{\emph{node.}}
\end{itemize}
The criteria of d-separation can be used to generate a causal graph
by calculating the various joint probability distributions over the
pairs of variables in a dataset. This is how the IC Algorithm\footnote{Algorithm definition adapted from \textcite{pearlCausalityModelsReasoning2000},
page 50.} generates a partial directed acyclic graph (PDAG, a graph where the
direction of some edges is ambivalent). 

\noindent\fbox{\begin{minipage}[t]{1\columnwidth - 2\fboxsep - 2\fboxrule}%
\textbf{IC Algorithm (Inductive Causation)}

\textbf{\textsc{Input}}: $\hat{P}$, a stable probability distribution
on a set of variables $V$.

\textbf{\textsc{Output}}: A PDAG $H(\hat{P})$, compatible with $\hat{P}$.
\begin{enumerate}
\item For each pair of variables $a$ and $b$ in $V,$ search for a set
$S_{ab}$ such that $(a\perp b\mid S_{ab})$ holds in $\hat{P}.$
This means $a$ and $b$ should be independent in $\hat{P}$, or d-separated,
when conditioned on $S_{ab}$. Create a DAG $G$ such that the vertices
$a$ and $b$ are connected with an edge if and only if no set $S_{ab}$
can be found.
\item For each pair of nonadjacent variables $a$ and $b$ with a common
neighbour $c$, check if $c\in S_{ab}$.
\begin{itemize}
\item If it is, then continue.
\item If not, then add directed edges pointing at $c$ i.e. $a\rightarrow c\leftarrow b$.
\end{itemize}
\item In the PDAG that results, orient as many of the undirected edges as
possible, subject to these two conditions:
\begin{itemize}
\item The orientation should not create a new $v$-structure.
\item The orientation should not create a directed cycle.
\end{itemize}
\end{enumerate}
\end{minipage}}

\,

Of course, there are other ways of discovering the causal graph (see
\textbf{Section VI(E)} for more details on the algorithms included
in AitiaExplorer). But the IC algorithm illustrates the close connection
between the causal model (understood as the joint probability distribution
in the system) and the causal graph.

\subsection{Motivation for AitiaExplorer}

The main problem with causal discovery is captured by Hyttinen et
al. (2016):
\begin{quote}
`` ... full knowledge of the true {[}causal{]} graph requires a rather
extensive understanding of the system under investigation. Data alone
is in general insufficient to uniquely determine the true causal graph.
Even complete discovery methods will usually leave the graph under
determined.''\footnote{\textcite{hyttinenDocalculusWhenTrue2016}.}
\end{quote}
Finding the causal graph of a system, or \textit{causal discovery,
}is\textit{ }difficult\textit{.} Even if your causal discovery method
creates an interesting graph, the graph may not be unique. This is
because the graph may be a member of a set of possibly Markov-equivalent
structures, each of which would satisfy the data.\footnote{See \textcite{jaberCausalIdentificationMarkov2018} for more details
on Markov-equivalent structures. }

However, there is some consolation:
\begin{quote}
``Algorithms that search for causal structure information ... do
not generally pin down all the possible causal details, but \textit{they
often provide important insight into the underlying causal relations}...''\footnote{\textcite{malinskyCausalDiscoveryAlgorithms2018}.}
(emphasis added).
\end{quote}
So this is an important motivation of AitiaExplorer and bolsters the
claim that AitiaExplorer enables exploratory causal analysis. AitiaExplorer
assists in the creation of causal graphs and can therefore provide
important insights into underlying causal structures.

This insight provides AitiaExplorer with a simple problem statement.

\noindent\fbox{\begin{minipage}[t]{1\columnwidth - 2\fboxsep - 2\fboxrule}%
\textbf{Problem Statement}

\textbf{\textsc{Input}}: A dataset with a large number of features
and with no known causal graph.

\textbf{\textsc{Task}}: To automatically select subsets of important
features from the dataset and create causal graphs candidates for
review based on these features. Then provide a metric to compare these
candidate causal graphs.%
\end{minipage}}

\section{Related Research}

\subsection{Causal Inference in Current Machine Learning Research}

There are many papers available in causal inference in machine learning
research as this is currently a popular topic for research. In order
to ascertain an idea of the amount of new research that is carried
out in this area, see the GitHub repository associated with \textcite{guoSurveyLearningCausality2018}.\footnote{https://github.com/rguo12/awesome-causality-algorithms}
For this reason, it is quite difficult to isolate the current ``state
of the art'' approach. However, each machine learning approach does
seem to have its own causal learning experiments, as one can see in
the selection of papers that are mentioned below. These papers were
interesting but differ from the approach that is undertaken in this
research:

Bengio et al.\footnote{\textcite{bengioMetaTransferObjectiveLearning2019}.}
train a deep learning network on labelled causal data and allow the
system to exploit small changes in probability distributions during
interventions to identify causal structures. As this method uses labelled
data, it does not fulfil the requirements of the current work and
will be left to one side.

Dasgupta et al.\footnote{\textcite{dasguptaCausalReasoningMetareinforcement2019}.}
uses a recurrent neural network with reinforcement learning in tasks
with different causal structures. As they note, their research is
``the first direct demonstration that causal reasoning can arise
out of model-free reinforcement learning''. They note ``traditional
formal approaches usually decouple the problems of causal induction
(inferring the structure of the underlying model from data), and causal
inference (estimating causal effects based on a known model)''. However
their work does not decouple these tasks (the use of \textit{causal
induction} here is what this document refers to as \textit{causal
discovery}). This use of reinforcement learning is highly attractive
as it is model free. However this approach will be left to one side
in this work in favour of less complex unsupervised methods.

Kalainathan et al.\footnote{\textcite{kalainathanStructuralAgnosticModeling2019}.}
present a new approach to causal learning called the Structural Agnostic
Model (SAM) which uses several different causal learning techniques
within a Generative Adversarial Neural network. They claim that this
provides a robust approach that has the advantages of multiple other
techniques combined. However, as this paper is concentrating on unsupervised
learning, the GAN approach will not be considered.

Bucur et al.\footnote{\textcite{bucurBayesianApproachInferring2018}.}
offer an interesting approach from genetic research. They attempt
to predict the causal structure of Gene Regulatory Networks (GRNs)
using the covariance values in the genetic data alongside existing
background knowledge of the genetic data priors to feed a Bayesian
algorithm. This research is illustrative of the wide applicability
of causal methods in many disciplines but it is too narrow in scope
to provide much assistance to the research in this document.

\subsection{Complementary Causal Inference Research}

There have been multiple papers published in the last few years in
learning causality that share some of the same objectives as this
work.

The paper from Pashami et al.\footnote{\textcite{pashamiCausalDiscoveryUsing2018a}.}
discusses ``the potential benefits, and explore{[}s{]} the hints
that clusters in the data can provide for causal discovery''. This
research provides some of the inspiration for \textit{AitiaExplorer},
in that unsupervised learning can provide, at the very least, some
heuristics for causal inference.

The work of Borboudakis and Tsamardinos\footnote{\textcite{borboudakisRobustVersatileCausal2016}.}
on \textit{ETIO} (from the Greek word for “cause”), a new ``general
causal discovery algorithm'', is in the same spirit as the research
outlined in this document. The authors create this tool for what they
call ``integrative causal discovery'' which is in keeping with the
pragmatic ensemble approach suggested for \textit{AitiaExplorer}.

Lin and Zhang\footnote{\textcite{linHowTackleExtremely2018}.} explore
the limitations of causal learning while also still retaining statistical
consistency. They outline a new learning theory that may provide some
ballast for a more general, ensemble approach to causal learning when
they say ``we should look for what can be achieved, and achieve the
best we can''.

\section{Requirements}

\subsection{Strategy}

The strategy behind AitiaExplorer is to create an exploratory causal
analysis (ECA) tool which will provide meaningful causal heuristics
in the causal analysis of a specific dataset.

This will satisfy the problem statement defined in Section 2.7.

\subsection{Primary Requirements}

The development of AitiaExplorer will be subject to the following
requirements, which follow on from the ECA discussion above:
\begin{itemize}
\item The software must place emphasis on augmenting human causal discovery
in a pragmatic way.
\item The software must be automated, at least partially.
\end{itemize}
These more technical requirements follow on from the problem statement:
\begin{itemize}
\item The software must be able to handle datasets with multiple features
and no known causal graph.
\item The software must be able to create multiple causal graphs and provide
a way to compare these causal graphs.
\end{itemize}
Due to the time scale and resources available for the development
of the software, the following will also apply:
\begin{itemize}
\item The software must make use of existing libraries and technology stacks.
\item The software must be open source.
\end{itemize}

\subsection{Secondary Requirements}

The development of AitiaExplorer will be subject to the following
requirements, which follow on from the primary requirements above:
\begin{itemize}
\item As the software needs to be at least partly automated, it follows
therefore that only unsupervised learning is an option. This means
no labelled data is required.
\item The software will need little or no data preparation work (beyond
the usual, such as scaling or encoding).
\item The software will not need specialist hardware such as GPUs (graphics
processing unit).
\item The software will take an ensemble approach, allowing multiple algorithms
to be tested in one pass.
\end{itemize}
For this reason, certain trends in current machine learning are not
pursued, as outlined above.

\section{Key Design Decisions}

\subsection{Programming Language Choice}

There are several probabilistic programming languages available and
these seemed like interesting choices for implementing a causal discovery
tool.

Several candidates were reviewed:
\begin{itemize}
\item Pyro - A universal probabilistic programming language written in Python.\footnote{https://pyro.ai}
\item Infer.NET - A .NET framework for probabilistic programming.\footnote{https://dotnet.github.io/infer/}
\item Gen - A general-purpose probabilistic programming system with programmable
inference.\footnote{https://probcomp.github.io/Gen/}
\end{itemize}
In the end, useful as these languages are for generic probabilistic
programming, none of the choices contained sufficient support for
learning causality. Most of the existing causal-related software is
implemented in Python (rather than in a subset of Python such as Pyro),
so it was deemed more appropriate for AitiaExplorer to be written
directly in Python.

\subsection{Causal Discovery Components}

After reviewing many causality-related software packages, several
were identified as being particularly useful for AitiaExplorer. 
\begin{itemize}
\item Amongst these, the Tetrad Project\footnote{http://www.phil.cmu.edu/tetrad/index.html }
appeared to be promising as this contains many causal discovery algorithms.
Unfortunately Tetrad is implemented in Java which it makes unsuitable
for AitiaExplorer. However, a Python package called py-causal\footnote{https://github.com/xunzheng/py-causal}
exposes these causal discovery algorithms via a Java-Python communication
layer. AitiaExplorer wraps the causal discovery algorithms provided
by py-causal.
\item The package CausalGraphicalModels\footnote{https://github.com/ijmbarr/causalgraphicalmodels}
was selected for displaying and manipulating causal graphs in Jupyter
Notebooks due to its simplicity and elegance.
\item The package pyAgrum\footnote{https://github.com/xunzheng/py-causal}
provides two algorithms used internally in AitiaExplorer:
\begin{itemize}
\item A target causal graph as supplied to AitiaExplorer represents the
full causal model contained within the dataset. Sometimes this is
known, but more often it is not. The greedy hill climbing algorithm
provided by pyAgrum is used to approximate causal graphs when no target
causal graph is provided. In theory, any causal discover algorithm
could be used as a benchmark in this case (see \textbf{Section VI}
for more details). However the algorithm used by pyAgrum is written
in C++ and is very fast which makes it an excellent choice.
\item The MIIC (Multivariate Information based Inductive Causation)\footnote{\textcite{vernyLearningCausalNetworks2017}.}
exposed by pyAgrum is used to identify latent unobserved edges in
causal graphs. There does not seem to be many implementations of this
algorithm (or similar) available in Python. Most implementations are
in R.
\end{itemize}
\end{itemize}

\subsection{Unsupervised Learning for Feature Selection and Extraction}

Although the approach using clustering as a causal heuristic in \textcite{pashamiCausalDiscoveryUsing2018a}
is very interesting, it is beyond the scope of this particular research
and does not provide a suitably pragmatic solution for AitiaExplorer.

Instead, two approaches using feature selection and extraction were
identified as being suitable for AitiaExplorer:
\begin{enumerate}
\item The use of clustering via Principal Feature Analysis to select important
features.
\item The use of standard supervised learning algorithms in an unsupervised
manner as outlined below in the Design and Implementation section.
These unsupervised versions of the algorithms can then be used for
feature selection.
\end{enumerate}

\subsection{Causal Graph Comparison Metrics}

In order to allow AitiaExplorer to compare causal graphs, AitiaExplorer
provides two metrics, Structural Hamming Distance (SHD)\footnote{\textcite{trabelsiBenchmarkingDynamicBayesian2013}.}
and Area Under the Precision-Recall Curve (AUPRC)\footnote{\textcite{boydAreaPrecisionRecallCurve2013}.}.
Both of these metrics are reasonably quick and an implementation of
each was available in Python. These measurements are provided in AitiaExplorer,
with an emphasis on SHD.

\section{Implementation}

\subsection{Design Overview}

A high level design for the AitiaExplorer system can be seen in the
\textbf{Figure 4} below. The system design is based loosely on the
design of the system outlined in Hyttinen et al. (2016).\footnote{\textcite{hyttinenDocalculusWhenTrue2016}.} 
\begin{itemize}
\item (A) The data to be analysed is passed in as a Pandas Dataframe. If
a target causal graph is supplied, this is used as a benchmark for
measuring all causal graphs generated by the system. If no target
causal graph is provided, the system assumes that it is unknown and
a target causal graph is generated from the data. This is then used
as a proxy benchmark by the system.
\item (B) The data is passed to the ensemble of unsupervised feature selection
algorithms.
\item (C) Each unsupervised feature selection algorithm selects what it
considers to be the most important features.
\item (D) The selected features from each unsupervised feature selection
algorithm are made available to the causal discovery algorithms.
\item (E) The data is passed to the ensemble of causal discovery algorithms.
\item (F) Each causal discovery algorithm is run in turn with just the selected
features from each unsupervised feature selection algorithm.
\item (G) A set of causal graphs and associated metrics is outputted.
\end{itemize}
\begin{figure}[H]
\includegraphics[scale=0.24]{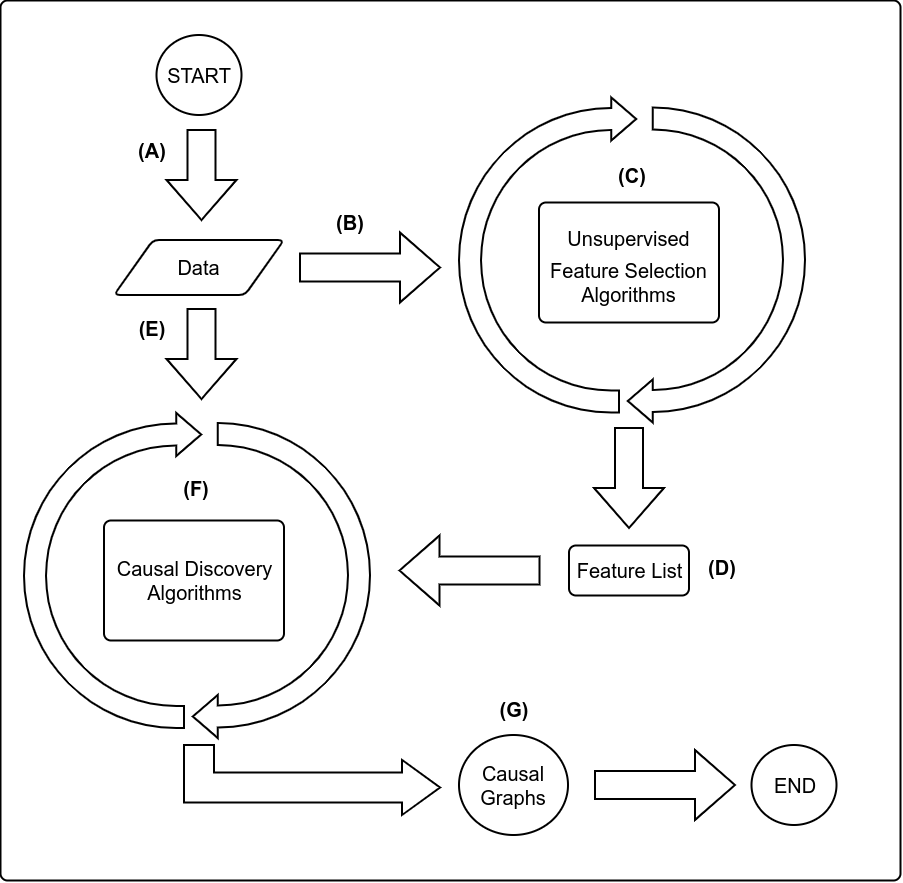}

\caption{Overall structure of the system.}
\end{figure}

The software is implemented as a set of Python classes and can be
used as either a Python library in an existing application, or as
an analysis tool within a Jupyter Notebook. 

\subsection{Principal Feature Analysis}

Principal Feature Analysis (PFA)\footnote{\textcite{luFeatureSelectionUsing2007}.}
is a method used for harnessing the dimensionality reduction ability
of Principal Component Analysis (PCA) whilst also being able to identify
the most important features that make up each principal component.
In PFA, the data is clustered using PCA and then the components are
fitted to a K-Means clustering algorithm. The most important features
are those with the minimum euclidean distance from a cluster centre. 

Principal Feature Analysis is available as one of the unsupervised
feature selection methods in AitiaExplorer.

\subsection{Turning Supervised Learning Algorithms into Unsupervised Learning
Algorithms}

Traditional supervised learning algorithms can be turned into unsupervised
learning algorithms\footnote{See \textcite{shiUnsupervisedLearningRandom2005} for more information.}
in the following way:
\begin{itemize}
\item Create suitable synthetic data from a reference distribution. In AitiaExplorer
this is achieved by using a Bayesian Gaussian Mixture Model (BGMM).
A BGMM can be used for clustering but it can also be used to model
the data distribution that best represents the data. This means that
a BGMM, when fitted to a specific dataset, can be used to provide
sample data, allowing the creation of synthetic data. 
\item This synthetic data can then be combined with real data, along with
an extra label that separates the synthetic data from the real data. 
\item This new dataset can then allow a classifier to be trained in an unsupervised
manner. 
\item The supervised learning algorithm learns to distinguish the original
data from the synthetic data.
\end{itemize}
The actual classification ability of the algorithms described above
is not vitally important for AitiaExplorer. Instead, AitiaExplorer
exploits the fact that as part of the training and classification
cycle above, each algorithm internally selects and orders features
in the dataset. AitiaExplorer puts the outputs of the classification
process to one side and instead queries each algorithm for the features
it considers to be the most important. 

There are other choices for creating plausible synthetic data that
would work e.g. using an autoencoder. However, the choice of BGMM
is pragmatic as befits the AitiaExplorer requirements outlined in
\textbf{Section IV}. The advantage of the BGMM is the actual scikit-learn
implementation. This implementation is fast, simple to use and actually
infers the effective number of clusters directly from the data. 

Of course, multiple tools and techniques exist that could could be
plugged into AitiaExplorer for feature selection. However, within
the constraints of this research, the Bayesian Gaussian Mixture Model
was found to be a good overall candidate that works well.

\subsection{Feature Selection Algorithms Available in AitiaExplorer}

The feature selection algorithms are listed in \textbf{Table II}.
PFA (Principal Feature Analysis) is implemented internally in AitiaExplorer,
XGBClassifier is provided as part of XGBoost\footnote{https://xgboost.ai/}
and the remainder of the algorithms are provided through SKLearn\footnote{https://scikit-learn.org/}.

\begin{table}
\begin{centering}
\begin{tabular}{|l|}
\hline 
\textbf{\scriptsize{}Algorithm}\tabularnewline
\hline 
\hline 
{\scriptsize{}Principal Feature Analysis (PFA)}\tabularnewline
\hline 
{\scriptsize{}XGBClassifier}\tabularnewline
\hline 
{\scriptsize{}Recursive Feature Elimination}\tabularnewline
\hline 
{\scriptsize{}Linear Regression}\tabularnewline
\hline 
{\scriptsize{}SGD Classifier}\tabularnewline
\hline 
{\scriptsize{}Random Forest Classifier}\tabularnewline
\hline 
{\scriptsize{}Gradient Boosting Classifier}\tabularnewline
\hline 
\end{tabular}
\par\end{centering}
\ 

\caption{Feature selection algorithms available in AitiaExplorer. }
\end{table}

\subsection{Causal Discovery Algorithms Available in AitiaExplorer}

There are two main types of causal discovery algorithms. These are
constraint-based algorithms and score based algorithms:
\begin{itemize}
\item Constraint-based algorithms build the graph by parsing the data and
looking for independent conditional probabilities as described in
\textbf{Section II}.
\item Score-based methods search over a set of possible graphs that fits
the data according to a metric.\footnote{For more information on this topic, see \textcite{triantafillouScorebasedVsConstraintbased2016}.}
\end{itemize}
AitiaExplorer allows the use of both types of causal discovery algorithms. 

\begin{table}
\begin{centering}
{\scriptsize{}}%
\begin{tabular}{|l|>{\raggedright}p{0.16\paperwidth}|}
\hline 
\textbf{\scriptsize{}Algorithm} & \textbf{\scriptsize{}Description}\tabularnewline
\hline 
\hline 
{\scriptsize{}bayesEst} & {\scriptsize{}Score-based - a revised Greedy Equivalence Search (GES)
algorithm. }\tablefootnote{{\scriptsize{}\textcite{chickeringOptimalStructureIdentification2002}.}}\tabularnewline
\hline 
{\scriptsize{}PC Algorithm} & {\scriptsize{}The original constraint-based algorithm.}\tablefootnote{{\scriptsize{}\textcite{spirtesCausationPredictionSearch2000}.}}\tabularnewline
\hline 
{\scriptsize{}FCI Algorithm} & {\scriptsize{}Constraint-based algorithm.}\tablefootnote{{\scriptsize{}\textcite{stroblApproximateKernelbasedConditional2017}.}}\tabularnewline
\hline 
{\scriptsize{}FGES Algorithm} & {\scriptsize{}Score-based - optimised and parallelised Greedy Equivalence
Search (GES).}\tablefootnote{{\scriptsize{}\textcite{meekCausalInferenceCausal1995}.}}\tabularnewline
\hline 
{\scriptsize{}GFCI Algorithm} & {\scriptsize{}Hybrid - a combination of the FGES and the FCI algorithm.}\tablefootnote{{\scriptsize{}\textcite{ogarrioHybridCausalSearch2016}.}}\tabularnewline
\hline 
{\scriptsize{}RFCI Algorithm} & {\scriptsize{}Constraint-based algorithm - a faster modification of
the FCI algorithm.}\tablefootnote{{\scriptsize{}\textcite{colomboLEARNINGHIGHDIMENSIONALDIRECTED2012}.}}\tabularnewline
\hline 
\end{tabular}{\scriptsize\par}
\par\end{centering}
\ 

\caption{Causal discovery algorithms available in AitiaExplorer.}
\end{table}

\section{Evaluation}

The evaluations below are based on the requirements outlined in \textbf{Section}
\textbf{IV}. Each scenario captures an illustration of how AitiaExplorer
meets these requirements. All of these evaluations were carried out
using AitiaExplorer running through a Jupyter Notebook.

\subsection{Evaluating the Unsupervised Learning Algorithms}

As explained earlier, AitiaExplorer uses supervised learning algorithms
in an unsupervised manner. A selection of these algorithms were trained
in this manner with data from the HEPAR II dataset\footnote{https://www.bnlearn.com/bnrepository/\#hepar2}
combined with synthetic data. The results are displayed in \textbf{Table
III}. Several of the classifiers have an almost perfect score on the
dataset in separating the real data from the synthetic data. Even
though the SGDClassifier does very poorly, it is still useful for
feature selection. LinearRegression and Principal Feature Analysis
have been omitted from this test as the score metric is not meaningful
for these algorithms.

\begin{table}[H]
\begin{centering}
{\scriptsize{}}%
\begin{tabular}{|l|r@{\extracolsep{0pt}.}l|}
\hline 
\textbf{\scriptsize{}Algorithm} & \multicolumn{2}{c|}{\textbf{\scriptsize{}Score}}\tabularnewline
\hline 
\hline 
{\scriptsize{}SGDClassifier} & {\scriptsize{}0}&{\scriptsize{}5004}\tabularnewline
\hline 
{\scriptsize{}RandomForestClassifier} & {\scriptsize{}1}&{\scriptsize{}0}\tabularnewline
\hline 
{\scriptsize{}GradientBoostingClassifier} & {\scriptsize{}1}&{\scriptsize{}0}\tabularnewline
\hline 
{\scriptsize{}XGBClassifier} & {\scriptsize{}1}&{\scriptsize{}0}\tabularnewline
\hline 
\end{tabular}{\scriptsize\par}
\par\end{centering}
\ 

\caption{Unsupervised learning algorithm scores.}
\end{table}

\subsection{Evaluating Causal Discovery When No Causal Graph is Supplied}

A target causal graph is important in AitiaExplorer because it offers
the user a simple way of comparing causal graphs produced by the system.
Each causal graph that is outputted has an Structural Hamming Distance
(SHD) and Area Under the Precision-Recall Curve (AUPRC) score given
against the target causal graph. When no target causal graph is supplied,
as outlined in \textbf{Section V(B)}, AitiaExplorer will generate
a proxy target graph using a fast greedy hill climbing algorithm.
The proxy target causal graph is not meant as a absolute measure of
correctness, but rather as an heuristic for the user to be able to
compare across causal graphs, independent of any specific combination
of causal discovery / feature selection algorithm. 

AitiaExplorer was run twice with data from the HEPAR II dataset. Both
runs used the same parameters, except that in the first run the known
target causal graph was supplied. In the second, no target graph was
supplied, forcing AitiaExplorer to create a proxy target causal graph.
The results from the first run with the known target causal graph
are displayed in \textbf{Table IV} overleaf. The results from the
second run with the proxy target causal graph are displayed in \textbf{Table
V} overleaf. As one would expect, the SHD scores are higher and the
AUPRC scores are lower in the second run. However, after a closer
inspection the changes in values are consistent across both runs.
The SHD values remain constant across all combinations in both runs.
Also, the AUPRC from the Random Forest Classifier fare worse than
other feature selection methods in both runs. These results verify
that the target causal graph, created by AitiaExplorer in run number
two, provides a reasonable proxy benchmark when no target causal graph
is supplied. The differences between the runs are proportional and
consistent.

\begin{table}
\begin{centering}
\begin{tabular}{|>{\raggedright}p{0.05\paperwidth}|>{\raggedright}p{0.15\paperwidth}|r@{\extracolsep{0pt}.}l|r@{\extracolsep{0pt}.}l|}
\hline 
\textbf{\scriptsize{}Causal Algorithm} & \textbf{\scriptsize{}Feature Selection Method} & \multicolumn{2}{c|}{\textbf{\scriptsize{}AUPRC}} & \multicolumn{2}{c|}{\textbf{\scriptsize{}SHD}}\tabularnewline
\hline 
\hline 
{\scriptsize{}PC} & {\scriptsize{}Linear Regression} & {\scriptsize{}0}&{\scriptsize{}5101} & \multicolumn{2}{c|}{{\scriptsize{}99}}\tabularnewline
\hline 
{\scriptsize{}FCI} & {\scriptsize{}Linear Regression} & {\scriptsize{}0}&{\scriptsize{}5101} & \multicolumn{2}{c|}{{\scriptsize{}99}}\tabularnewline
\hline 
{\scriptsize{}RFCI} & {\scriptsize{}Linear Regression} & {\scriptsize{}0}&{\scriptsize{}5101} & \multicolumn{2}{c|}{{\scriptsize{}99}}\tabularnewline
\hline 
{\scriptsize{}PC} & {\scriptsize{}Random Forest Classifier} & {\scriptsize{}0}&{\scriptsize{}26505} & \multicolumn{2}{c|}{{\scriptsize{}99}}\tabularnewline
\hline 
{\scriptsize{}FCI} & {\scriptsize{}Random Forest Classifier} & {\scriptsize{}0}&{\scriptsize{}26505} & \multicolumn{2}{c|}{{\scriptsize{}99}}\tabularnewline
\hline 
{\scriptsize{}RFCI} & {\scriptsize{}Random Forest Classifier} & {\scriptsize{}0}&{\scriptsize{}26505} & \multicolumn{2}{c|}{{\scriptsize{}99}}\tabularnewline
\hline 
{\scriptsize{}PC} & {\scriptsize{}Recursive Feature Elimination} & {\scriptsize{}0}&{\scriptsize{}5101} & \multicolumn{2}{c|}{{\scriptsize{}99}}\tabularnewline
\hline 
{\scriptsize{}FCI} & {\scriptsize{}Recursive Feature Elimination} & {\scriptsize{}0}&{\scriptsize{}5101} & \multicolumn{2}{c|}{{\scriptsize{}99}}\tabularnewline
\hline 
{\scriptsize{}RFCI} & {\scriptsize{}Recursive Feature Elimination} & {\scriptsize{}0}&{\scriptsize{}5101} & \multicolumn{2}{c|}{{\scriptsize{}99}}\tabularnewline
\hline 
\end{tabular}
\par\end{centering}
\ 

\caption{Causal discovery when causal graph is supplied.}
\end{table}

\begin{table}
\begin{centering}
\begin{tabular}{|>{\raggedright}p{0.05\paperwidth}|>{\raggedright}p{0.15\paperwidth}|r@{\extracolsep{0pt}.}l|r@{\extracolsep{0pt}.}l|}
\hline 
\textbf{\scriptsize{}Causal Algorithm} & \textbf{\scriptsize{}Feature Selection Method} & \multicolumn{2}{c|}{\textbf{\scriptsize{}AUPRC}} & \multicolumn{2}{c|}{\textbf{\scriptsize{}SHD}}\tabularnewline
\hline 
\hline 
{\scriptsize{}PC} & {\scriptsize{}Linear Regression} & {\scriptsize{}0}&{\scriptsize{}51255} & \multicolumn{2}{c|}{{\scriptsize{}123}}\tabularnewline
\hline 
{\scriptsize{}FCI} & {\scriptsize{}Linear Regression} & {\scriptsize{}0}&{\scriptsize{}51255} & \multicolumn{2}{c|}{{\scriptsize{}123}}\tabularnewline
\hline 
{\scriptsize{}RFCI} & {\scriptsize{}Linear Regression} & {\scriptsize{}0}&{\scriptsize{}51255} & \multicolumn{2}{c|}{{\scriptsize{}123}}\tabularnewline
\hline 
{\scriptsize{}PC} & {\scriptsize{}Random Forest Classifier} & {\scriptsize{}0}&{\scriptsize{}01255} & \multicolumn{2}{c|}{{\scriptsize{}125}}\tabularnewline
\hline 
{\scriptsize{}FCI} & {\scriptsize{}Random Forest Classifier} & {\scriptsize{}0}&{\scriptsize{}01255} & \multicolumn{2}{c|}{{\scriptsize{}125}}\tabularnewline
\hline 
{\scriptsize{}RFCI} & {\scriptsize{}Random Forest Classifier} & {\scriptsize{}0}&{\scriptsize{}01255} & \multicolumn{2}{c|}{{\scriptsize{}125}}\tabularnewline
\hline 
{\scriptsize{}PC} & {\scriptsize{}Recursive Feature Elimination} & {\scriptsize{}0}&{\scriptsize{}51255} & \multicolumn{2}{c|}{{\scriptsize{}123}}\tabularnewline
\hline 
{\scriptsize{}FCI} & {\scriptsize{}Recursive Feature Elimination} & {\scriptsize{}0}&{\scriptsize{}51255} & \multicolumn{2}{c|}{{\scriptsize{}123}}\tabularnewline
\hline 
{\scriptsize{}RFCI} & {\scriptsize{}Recursive Feature Elimination} & {\scriptsize{}0}&{\scriptsize{}51255} & \multicolumn{2}{c|}{{\scriptsize{}123}}\tabularnewline
\hline 
\end{tabular}
\par\end{centering}
\ 

\caption{Causal discovery when no causal graph is supplied.}
\end{table}

\subsection{Evaluating Causal Discovery with a Set Number of Features}

AitiaExplorer allows one to select the number of features that are
selected in the causal discovery process. In this example, AitiaExplorer
was run with a combination of causal discovery / feature selection
algorithms. For clarity and simplicity, a small simulated dataset
was used and only 7 features of a possible 10 features were selected.
The results from the run are displayed in \textbf{Table VI} below.
The best results have a lower SHD and a higher AUPRC.

\begin{table}
\begin{centering}
\begin{tabular}{|>{\raggedright}p{0.03\paperwidth}|>{\raggedright}p{0.05\paperwidth}|>{\raggedright}p{0.1\paperwidth}|r@{\extracolsep{0pt}.}l|r@{\extracolsep{0pt}.}l|}
\hline 
\textbf{\scriptsize{}Run Index} & \textbf{\scriptsize{}Causal Algorithm} & \textbf{\scriptsize{}Feature Selection Method} & \multicolumn{2}{c|}{\textbf{\scriptsize{}AUPRC}} & \multicolumn{2}{c|}{\textbf{\scriptsize{}SHD}}\tabularnewline
\hline 
\hline 
{\scriptsize{}0} & {\scriptsize{}PC} & {\scriptsize{}Linear Regression} & {\scriptsize{}0}&{\scriptsize{}662500} & \multicolumn{2}{c|}{{\scriptsize{}8}}\tabularnewline
\hline 
{\scriptsize{}1} & {\scriptsize{}FCI} & {\scriptsize{}Linear Regression} & {\scriptsize{}0}&{\scriptsize{}662500} & \multicolumn{2}{c|}{{\scriptsize{}8}}\tabularnewline
\hline 
{\scriptsize{}2} & {\scriptsize{}FGES} & {\scriptsize{}Linear Regression} & {\scriptsize{}0}&{\scriptsize{}662500} & \multicolumn{2}{c|}{{\scriptsize{}8}}\tabularnewline
\hline 
{\scriptsize{}3} & {\scriptsize{}GFCI} & {\scriptsize{}Linear Regression} & {\scriptsize{}0}&{\scriptsize{}662500} & \multicolumn{2}{c|}{{\scriptsize{}8}}\tabularnewline
\hline 
{\scriptsize{}4} & {\scriptsize{}RFCI} & {\scriptsize{}Linear Regression} & {\scriptsize{}0}&{\scriptsize{}662500} & \multicolumn{2}{c|}{{\scriptsize{}8}}\tabularnewline
\hline 
{\scriptsize{}5} & {\scriptsize{}PC} & {\scriptsize{}Principal Feature Analysis} & {\scriptsize{}0}&{\scriptsize{}370313} & \multicolumn{2}{c|}{{\scriptsize{}10}}\tabularnewline
\hline 
{\scriptsize{}6} & {\scriptsize{}FCI} & {\scriptsize{}Principal Feature Analysis} & {\scriptsize{}0}&{\scriptsize{}370313} & \multicolumn{2}{c|}{{\scriptsize{}10}}\tabularnewline
\hline 
{\scriptsize{}7} & {\scriptsize{}FGES} & {\scriptsize{}Principal Feature Analysis} & {\scriptsize{}0}&{\scriptsize{}370313} & \multicolumn{2}{c|}{{\scriptsize{}10}}\tabularnewline
\hline 
{\scriptsize{}8} & {\scriptsize{}GFCI} & {\scriptsize{}Principal Feature Analysis} & {\scriptsize{}0}&{\scriptsize{}370313} & \multicolumn{2}{c|}{{\scriptsize{}10}}\tabularnewline
\hline 
{\scriptsize{}9} & {\scriptsize{}RFCI} & {\scriptsize{}Principal Feature Analysis} & {\scriptsize{}0}&{\scriptsize{}370313} & \multicolumn{2}{c|}{{\scriptsize{}10}}\tabularnewline
\hline 
{\scriptsize{}10} & {\scriptsize{}PC} & {\scriptsize{}Random Forest} & {\scriptsize{}0}&{\scriptsize{}495833} & \multicolumn{2}{c|}{{\scriptsize{}9}}\tabularnewline
\hline 
{\scriptsize{}11} & {\scriptsize{}FCI} & {\scriptsize{}Random Forest} & {\scriptsize{}0}&{\scriptsize{}370313} & \multicolumn{2}{c|}{{\scriptsize{}10}}\tabularnewline
\hline 
{\scriptsize{}12} & {\scriptsize{}FGES} & {\scriptsize{}Random Forest} & {\scriptsize{}0}&{\scriptsize{}370313} & \multicolumn{2}{c|}{{\scriptsize{}10}}\tabularnewline
\hline 
{\scriptsize{}13} & {\scriptsize{}GFCI} & {\scriptsize{}Random Forest} & {\scriptsize{}0}&{\scriptsize{}370313} & \multicolumn{2}{c|}{{\scriptsize{}10}}\tabularnewline
\hline 
{\scriptsize{}14} & {\scriptsize{}RFCI} & {\scriptsize{}Random Forest} & {\scriptsize{}0}&{\scriptsize{}370313} & \multicolumn{2}{c|}{{\scriptsize{}10}}\tabularnewline
\hline 
{\scriptsize{}15} & {\scriptsize{}PC} & {\scriptsize{}Recursive Feature Elimination} & {\scriptsize{}0}&{\scriptsize{}286979} & \multicolumn{2}{c|}{{\scriptsize{}11}}\tabularnewline
\hline 
{\scriptsize{}16} & {\scriptsize{}FCI} & {\scriptsize{}Recursive Feature Elimination} & {\scriptsize{}0}&{\scriptsize{}078125} & \multicolumn{2}{c|}{{\scriptsize{}12}}\tabularnewline
\hline 
{\scriptsize{}17} & {\scriptsize{}FGES} & {\scriptsize{}Recursive Feature Elimination} & {\scriptsize{}0}&{\scriptsize{}078125} & \multicolumn{2}{c|}{{\scriptsize{}12}}\tabularnewline
\hline 
{\scriptsize{}18} & {\scriptsize{}GFCI} & {\scriptsize{}Recursive Feature Elimination} & {\scriptsize{}0}&{\scriptsize{}078125} & \multicolumn{2}{c|}{{\scriptsize{}12}}\tabularnewline
\hline 
{\scriptsize{}19} & {\scriptsize{}RFCI} & {\scriptsize{}Recursive Feature Elimination} & {\scriptsize{}0}&{\scriptsize{}078125} & \multicolumn{2}{c|}{{\scriptsize{}12}}\tabularnewline
\hline 
{\scriptsize{}20} & {\scriptsize{}PC} & {\scriptsize{}XGBoost} & {\scriptsize{}0}&{\scriptsize{}286979} & \multicolumn{2}{c|}{{\scriptsize{}11}}\tabularnewline
\hline 
{\scriptsize{}21} & {\scriptsize{}FCI} & {\scriptsize{}XGBoost} & {\scriptsize{}0}&{\scriptsize{}286979} & \multicolumn{2}{c|}{{\scriptsize{}11}}\tabularnewline
\hline 
{\scriptsize{}22} & {\scriptsize{}FGES} & {\scriptsize{}XGBoost} & {\scriptsize{}0}&{\scriptsize{}286979} & \multicolumn{2}{c|}{{\scriptsize{}11}}\tabularnewline
\hline 
{\scriptsize{}23} & {\scriptsize{}GFCI} & {\scriptsize{}XGBoost} & {\scriptsize{}0}&{\scriptsize{}286979} & \multicolumn{2}{c|}{{\scriptsize{}11}}\tabularnewline
\hline 
{\scriptsize{}24} & {\scriptsize{}RFCI} & {\scriptsize{}XGBoost} & {\scriptsize{}0}&{\scriptsize{}286979} & \multicolumn{2}{c|}{{\scriptsize{}11}}\tabularnewline
\hline 
\end{tabular}
\par\end{centering}
\ 

\caption{Causal discovery with a set number of features.}
\end{table}

One can in the plot the relationship between the SHD and the AUPRC
as in \textbf{Figure 5} above. In general, the lower SHD is associated
with a higher AUPRC, suggesting that the combination of causal discovery
/ feature selection algorithms from earlier runs may be optimal. These
results demonstrate a simple use case of AitiaExplorer which allows
the user to fix the number of features and test multiple combinations
of algorithms in one step. These combinations can then be compared
for further exploration.

\begin{figure}[t]
\includegraphics[scale=0.4]{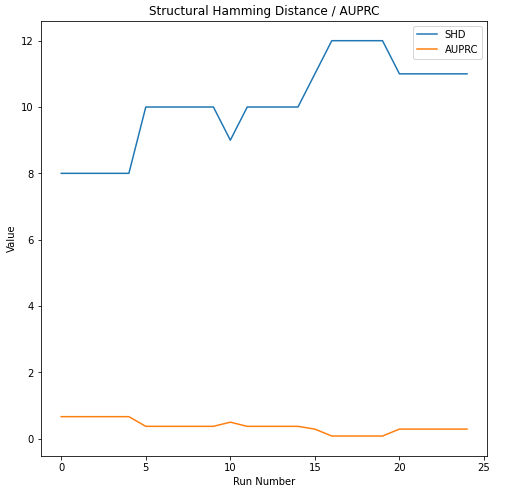}

\caption{Causal discovery with a set number of features.}
\end{figure}

\subsection{Evaluating Causal Discovery Within a Range of Features}

AitiaExplorer allows one to select a numeric range of features that
are selected in the causal discovery process. In this example, AitiaExplorer
was run with a range of between 10 and 20 features from the HEPAR
II dataset. The PC and FGES causal discovery algorithms and the Linear
Regression and Random Forest feature selection algorithms were selected.
AitiaExplorer then ran the selected combinations of algorithms across
the data for each number of features, from 10 to 20. The results for
the SHD and AUPRC values are graphed below in \textbf{Figure 6} and
\textbf{Figure 7}. 

\begin{figure}[H]
\includegraphics[scale=0.36]{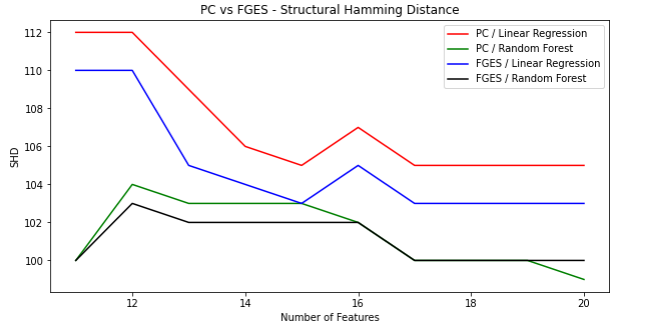}

\caption{SHD - causal discovery with a range of features.}
\end{figure}

The results of this example are interesting because here a score-based
causal discovery algorithm (PC) is pitted against a constraint-based
causal discovery algorithm. Both causal discovery algorithms have
a lower SHD when the number of features goes above 17. However, it
appears that the AUPRC of the PC causal discovery algorithm is higher
when used with the Random Forest Classifier. This knowledge can be
used in further causal analysis. This example is an indication of
how AitiaExplorer automatically provides a way of comparing multiple
methods of producing a causal graph to see which is the most promising.

\begin{figure}[H]
\includegraphics[scale=0.36]{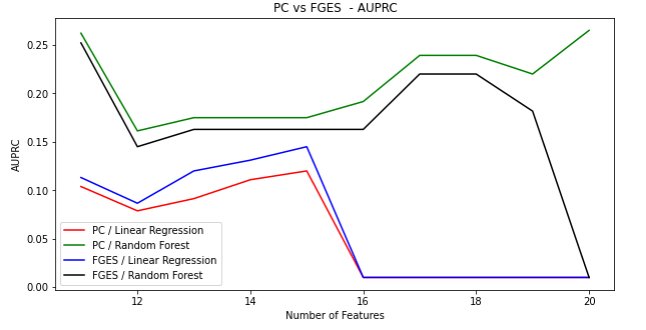}

\caption{AUPRC - causal discovery with a range of features.}
\end{figure}

\subsection{Evaluating Causal Discovery with Latent Unobserved Variables}

The nodes in a causal graph are the observable or measured causal
features. However, unobserved or latent variables can be inferred
by certain causal discovery algorithms\footnote{For instance, see the usage of FCI in \cite{shenChallengesOpportunitiesCausal2020}.
However, finding latent variables ``is a nontrivial task that is
still an active area of research'' according to \textcite{petersElementsCausalInference2017},
page 184, and thus beyond the scope of the current research.}. These confounding variables (where the node is an unobservable common
cause of two observable nodes) can be represented on a causal graph
as an empty node as in \textbf{Figure 8}. In this example, the latent
variable could be $hosepipe$ which could be a cause of both $ice$
and $wet$ depending on the season. 

\begin{figure}[h]
\begin{centering}
\includegraphics[scale=0.33]{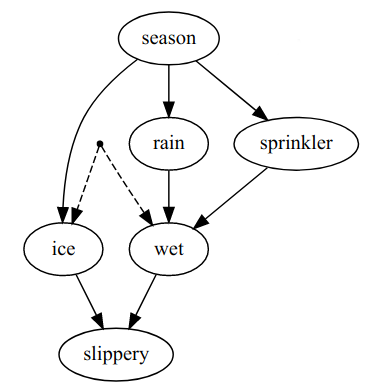}
\par\end{centering}
\caption{Example of a causal graph with latent edges.}
\end{figure}

AitiaExplorer will identify latent unobserved features in the causal
discovery process as outlined in \textbf{Section V(B)}. In this example,
AitiaExplorer was run with a selection of 30 features from the HEPAR
II dataset. Once the system has finished running the causal discovery
process, once can display the causal graphs that contain latent unobserved
features. A section of such a causal graph is displayed in \textbf{Figure
9 }above (highlight added). One can also get a list of the edges where
these missing nodes are. In this case, these edges are identified
as \texttt{{[}('Cirrhosis', 'platelet'), ('Cirrhosis', 'alcoholism'){]}}.
These values can then be used in further analysis, as required.

\begin{figure}[H]
\includegraphics[scale=0.2]{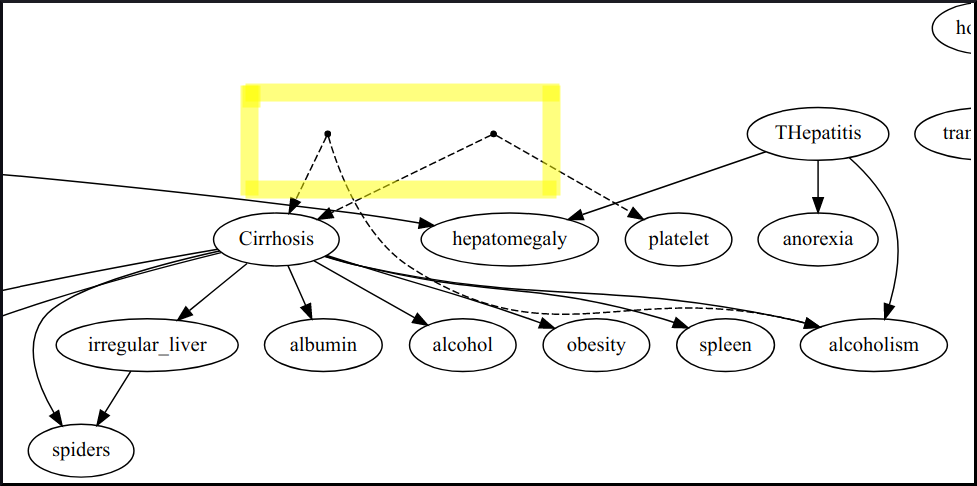}

\caption{A causal graph with latent unobserved edges.}
\end{figure}

\subsection{Achievement of Requirements}

AitiaExplorer can now be evaluated as to whether it meets the primary
and secondary requirements as outlined in \textbf{Section IV}. 

\subsubsection{Primary Requirements}
\begin{itemize}
\item \textit{The software must place emphasis on augmenting human causal
discovery in a pragmatic way:} AitiaExplorer meets this requirement.
The examples in this section illustrate how AitiaExplorer is an exploratory
causal analysis, extending the ability of a user to explore the causal
structures inherent in a given dataset. 
\item \textit{The software must be automated, at least partially:} AitiaExplorer
meets this requirement. As shown in \textbf{Sections VII(B)}, \textbf{VII(C)}
and \textbf{VII(D)}, AitiaExplorer just needs the user to pass in
the required parameters and the system will do the rest.
\item \textit{The software must be able to handle datasets with multiple
features and no known causal graph:} AitiaExplorer meets this requirement.
Several of the examples here use the HEPAR II dataset which contains
ten thousand records and seventy feature nodes. 
\item \textit{The software must be able to create multiple causal graphs
and provide a way to compare these causal graphs:} AitiaExplorer meets
this requirement. As shown in \textbf{Sections VII(C)} and \textbf{VII(D)},
AitiaExplorer provides the SHD and AUPRC metrics for each combination
of algorithm. These values are returned in a dataframe and can be
plotted using standard Python tools in a Jupyter Notebook.
\item \textit{The software must make use of existing libraries and technology
stacks:} AitiaExplorer meets this requirement. AitiaExplorer relies
upon several open source Python frameworks as outlined in \textbf{Section
V}.
\item \textit{The software must be open source:} AitiaExplorer meets this
requirement. The source code is available on GitHub under a permissive
open source license.
\end{itemize}

\subsubsection{Secondary Requirements}
\begin{itemize}
\item \textit{As the software needs to be at least partly automated, it
follows therefore, that only unsupervised learning is an option:}
AitiaExplorer meets this requirement. All feature selection algorithms
used are unsupervised. No labelled training data is required.
\item \textit{The software will need little or no data preparation work:}
AitiaExplorer meets this requirement. AitiaExplorer accepts a dataset
in a standard Pandas Dataframe.
\item \textit{The software will not need specialist hardware such as GPUs:}
AitiaExplorer meets this requirement. All evaluations were run on
a laptop without the use of a GPU and most were completed in under
an hour.
\item \textit{The software will take an ensemble approach, allowing multiple
algorithms to be tested in one pass:} AitiaExplorer meets this requirement.
AitiaExplorer will take an arbitrary number of causal discovery and
feature selection algorithms, once they are defined with the system.
\end{itemize}

\section{Conclusion}

AitiaExplorer provides an efficient solution to the problem statement
from Section 2.7. The software is a useful exploratory causal analysis
tool that automatically selects subsets of important features from
a dataset and creates causal graph candidates for review based on
these features. A metric is also provided to compare these candidate
causal graphs.

\subsection{SWOT Analysis}

\subsubsection{Strengths}
\begin{itemize}
\item AitiaExplorer met the requirements for an exploratory causal analysis
as set out in \textbf{Section IV}.
\item AitiaExplorer demonstrates that one can build a system that augments
the ability of a user to find candidate causal graphs efficiently.
\item The Python language and ecosystem is an excellent choice for this
kind of project. The availability of many excellent causality-related
libraries is a major advantage. Working within a Jupyter Notebook
with AitiaExplorer is very straightforward and productive.
\end{itemize}

\subsubsection{Weaknesses}
\begin{itemize}
\item Causal discovery is still very difficult without an extensive understanding
of the system under investigation. Many of the causal graphs returned
by AitiaExplorer are poor candidates. This is part of the challenge
of causal discovery.
\item The SHD and AUPRC metrics as provided by AitiaExplorer are useful,
but only provide a shallow comparison metric between graphs. With
more time and resources, better comparison metrics with more detailed
analysis could be provided.
\end{itemize}

\subsubsection{Opportunities}
\begin{itemize}
\item AitiaExplorer could be extended in several interesting ways, perhaps
to allow further analysis of causal models. As per the terminology
outlined in \textbf{Section II(F)}, AitiaExplorer is primarily a causal
discovery tool and works on the level of causal graphs. With more
time and resources, AitiaExplorer\textit{ }could become a causal inference
tool also, and allow analysis of the underlying causal models behind
the causal graphs. 
\item A version of AitiaExplorer with causal feature selection methods alongside
classical feature selection methods for comparison would be a very
interesting and potentially useful research project.
\end{itemize}

\subsubsection{Threats}
\begin{itemize}
\item AitiaExplorer was tested against reasonably large datasets and worked
efficiently. However, some of the datasets where exploratory causal
analysis would be really useful are very large indeed, often with
thousands of features and hundreds of thousands of records. It is
unknown how well AitiaExplorer would perform under this type of workload.
\end{itemize}

\subsection{Future Work}

It was hoped that AitiaExplorer could be extended to include support
for the DoWhy calculus\footnote{\textcite{pearlBookWhyNew2019}} of
Judea Pearl which would open up the ability for AitiaExplorer to explore
counterfactual causal graphs and perhaps compare scenarios of several
candidate graphs. It was also desirable to include some different,
more innovative algorithms in AitiaExplorer, such as the NOTEARS algorithm\footnote{\textcite{zhengDAGsNOTEARS2018}.}
and the Boruta algorithm\footnote{\textcite{kursaFeatureSelectionBoruta2010}}.
However, due to time and resource constraints, this interesting work
will have to be carried out at a future time. 

\subsection{Acknowledgements}

The author would like to thank Dr. Alessio Benavoli and Dr Pepijn
van de Ven for their professional advice and support.

\pagebreak{}

\printbibliography[heading=bibintoc]

\end{document}